\documentclass[runningheads]{llncs}
\usepackage{times}
\usepackage{graphicx}
\usepackage{latexsym}
\usepackage{times}  
\usepackage{helvet} 
\usepackage{courier}  
\usepackage[hyphens]{url}  
\usepackage{graphicx}  
\usepackage{hyperref}
\usepackage{cite}
\usepackage{amsmath,amssymb,amsfonts}
\usepackage{algorithmic}
\usepackage{subfig}
\usepackage{textcomp}
\usepackage{fancyvrb}
\usepackage{xcolor}
\usepackage{multirow}
\usepackage{subfig}
\usepackage[normalem]{ulem}
\useunder{\uline}{\ul}{}
\def\BibTeX{{\rm B\kern-.05em{\sc i\kern-.025em b}\kern-.08em
    T\kern-.1667em\lower.7ex\hbox{E}\kern-.125emX}}

\begin{document}

\title{Self-Supervised Log Parsing\\
}

\newcommand{\jasmin}[1]{{\color{green} Jasmin: #1}}
\newcommand{\sasho}[1]{{\color{red} Sasho: #1}}
\newcommand{\alex}[1]{{\color{blue} Alex: #1}}
\newcommand{\archiv}[1]{{\color{purple} Archiv: #1}}

\author{Sasho Nedelkoski\inst{1,3}\and
Jasmin Bogatinovski\inst{1,3}\and
Alexander Acker\inst{1}\and
Jorge Cardoso\inst{2}\and
Odej Kao\inst{1}}
\authorrunning{S. Nedelkoski et al.}
%
\institute{Distributed Systems, 
TU Berlin, Berlin, Germany \\
\email{nedelkoski, jasmin.bogatinovski, alexander.acker, odej.kao@tu-berlin.de}\\
\and
Department of Informatics Engineering/CISUC, University of Coimbra, Portugal\\
\email{jcardoso@dei.uc.pt}\\
\and 
Equal contribution
}
\maketitle






\begin{abstract}
Logs are extensively used during the development and maintenance of software systems. They collect runtime events and allow tracking of code execution, which enables a variety of critical tasks such as troubleshooting and fault detection. However, large-scale software systems generate massive volumes of semi-structured log records, posing a major challenge for automated analysis. 
Parsing semi-structured records with free-form text log messages into structured templates is the first and crucial step that enables further analysis. Existing approaches rely on log-specific heuristics or manual rule extraction. These are often specialized in parsing certain log types, and thus, limit performance scores and generalization. We propose a novel parsing technique called \texttt{NuLog} that utilizes a self-supervised learning model and formulates the parsing task as masked language modeling (MLM). In the process of parsing, the model extracts summarizations from the logs in the form of a vector embedding. This allows the coupling of the MLM as pre-training with a downstream anomaly detection task. 
We evaluate the parsing performance of \texttt{NuLog} on 10 real-world log datasets and compare the results with 12 parsing techniques. The results show that \texttt{NuLog} outperforms existing methods in parsing accuracy with an average of 99\% and achieves the lowest edit distance to the ground truth templates. Additionally, two case studies are conducted to demonstrate the ability of the approach for log-based anomaly detection in both supervised and unsupervised scenario. The results show that \texttt{NuLog} can be successfully used to support troubleshooting tasks. 

The implementation is available at \href{https://github.com/nulog/nulog}{\texttt{https://github.com/nulog/nulog}}.
\end{abstract}


\section{Introduction} \label{introduction}

Current IT systems are a combination of complex multi-layered software and hardware. They enable applications of ever-increasing complexity and system diversity, where many technologies such as the Internet of Things (IoT), distributed processing frameworks, databases, and operating systems are used. The complexity and diversity of the systems relate to high managing and maintenance overhead for the operators to a point where they are no longer sufficient to holistically operate and manage these systems. 
Therefore, service providers are deploying various measures by introducing additional AI solutions for
anomaly detection, error analysis, and recovery to the IT ecosystem~\cite{8814585}. 
The foundation for these data-driven troubleshooting solutions is the availability of data that describe the state of the systems. 
The large variety of technologies leads to diverse data compelling the developed methods to generalize well over different applications, operating systems, or cloud infrastructure management tools.

One specific data source -- the logs, are commonly used to inspect the behavior of an IT system. They represent interactions between data, files, services, or applications, which are typically utilized by the developers, DevOps teams, and AI methods to understand system behaviors to detect, localize, and resolve problems that may arise~\cite{8752866}. 
The first step for understanding log information and their utilization for further automated analysis is to parse them. The content of a log record is an unstructured free-text written by software developers, which makes it difficult to structure. It is a composition of constant string templates and variable values. The template is the logging instruction (e.g. \textit{print()}, \textit{log.info()}) from which the log message is produced. It records a specific system event. The general objective of a log parser is the transformation of the unstructured free-text into a structured log template and an associated list of variables. For example, the template \textit{"Attempting claim: memory $\langle * \rangle$ MB, disk $\langle * \rangle$ GB, vcpus $\langle * \rangle$ CPU"} is associated with the variable list \textit{["2048", "20, "1"]}. Here, $\langle * \rangle$ denotes the position of each variable and is associated with the positions of the values within the list. The variable list can be empty if a template does not contain variable parts.

Traditional log parsing techniques rely on regular expressions designed and maintained by human experts. Large systems consisting of diverse software and hardware components render it intricate to maintain this manual effort. Additionally, frequent software updates necessitate constant checking and adjusting of these statements, which is a tedious and error-prone task. Related log parsing methods~\cite{he2017drain, du2016spell, Hamooni2016LogMineFP, zhu2017deep} depend on parse trees, heuristics, and domain knowledge. They are either specialized to perform well on logs from specific systems or can reliably parse data with a low variety of unique templates. Analyzing the performance of existing log parsing methods on a variety of diverse systems reveals their lack of robustness to produce consistently good parsing results. This implies the necessity to choose a parsing method for the application or system at hand and incorporating domain-specific knowledge. Operators of large IT infrastructures would end up with the overhead of managing different parsing methods for their components whereof each need to be accordingly understood. Based on this, we state that log parsing methods have to be accurate on log data from various systems ranging from single applications over mobile operating systems to cloud infrastructure management platforms with the least human intervention.

\textbf{Contribution.} 
We propose a self-supervised method for log parsing \texttt{NuLog}, which utilizes the transformer architecture~\cite{vaswani2017attention, devlin2018bert}. Self-supervised learning is a form of unsupervised learning where parts of the data provide supervision. To build the model, the learning task is formulated such that the presence of a word on a particular position in a log message is conditioned on its context.
The key idea for parsing is that the correct prediction of the masked word means that the word is a part of the log template otherwise, it is a parameter of the log. The advantages of this approach are that it can produce both a log template and a numerical vector sumarization, while domain knowledge is not needed.
Through exhaustive experimentation, we show that \texttt{NuLog} outperforms the previous state of the art log parsing methods and achieves the best scores overall. The model is robust and generalizes well across different datasets.
Further, we illustrate two use cases, supervised and unsupervised, on how the model can be coupled with and fine-tuned for downstream tasks like anomaly detection. The results suggest that the knowledge obtained during the masked language modeling in for the log parsing phase is useful as a good prior knowledge for the downstream tasks.

\section{Related Work}\label{relatedwork}

\begin{figure}[!b]
\centerline{\includegraphics[scale=0.38]{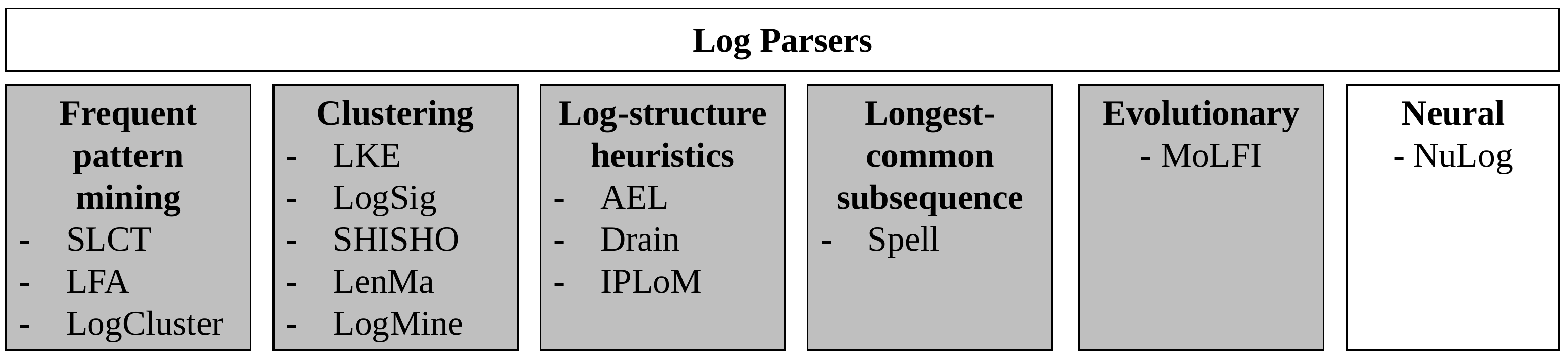}}
\caption{Taxonomy of log parses according to underlying technology they adopt.}
\label{fig:1}
\end{figure}

Automated log parsing is important due to its practical relevance for the maintenance and troubleshooting of software systems. A significant amount of research and development for automated log parsing methods has been published in both industry and academia~\cite{zhu2019tools, he2016evaluation}. Parsing techniques can be distinguished in various aspects, including technological, operation mode, and preprocessing. In \figurename~\ref{fig:1}, we give an overview of the existing methods.

\textbf{Clustering} The main assumption in these methods is that the message types coincide in similar groups. Various clustering methods with proper string matching distances have been used. LKE~\cite{fu2009execution} applies weighted edit distance with hierarchical clustering to do log key extraction and a group splitting strategy to fine-tune the obtained log groups. LogSig~\cite{tang2011logsig} is a message signature-based algorithm that searches for the most representative message signatures, heavily utilizing domain knowledge to determine the number of clusters. SHISO~\cite{mizutani2013incremental} is creating a structured tree using the nodes generated from log messages which enables a real-time update of new log messages if a match with previously existing log templates fails. LenMa~\cite{shima2016length} utilizes a clustering approach based on sequences of word lengths appearing in the logs. LogMine~\cite{Hamooni2016LogMineFP} creates a hierarchy of log templates, that allows the user to choose the description level of interest.

\textbf{Frequent pattern mining} assumes that a message type is a frequent set of tokens that appear throughout the logs. The procedures involve creating frequent sets, grouping the log messages, and extraction of message types. Representative parsers for this group are SLCT, LFA, and LogCluster~\cite{xu2009detecting, nagappan2010abstracting, nandi2016anomaly}.

\textbf{Evolutionary} is the last category. Its member MoLFI~\cite{messaoudi2018search} uses an evolutionary approach to find the Pareto optimal set of message templates.

\textbf{Log-structure heuristics methods} produce the best results among the different adopted techniques~\cite{zhu2019tools,he2016evaluation}. They usually exploit different properties that emerge from the structure of the log. The state-of-the-art Drain~\cite{he2017drain} assumes that at the beginning of the logs the words do not vary too much. It uses this assumption to create a tree of fixed depth which can be easily modified for new groups. Other parsing methods in this group are IPLoM and AEL~\cite{xu2009detecting, jiang2008automated}

\textbf{Longest-common sub-sequence} uses the longest common subsequence algorithm to dynamically extract log patterns from incoming logs. Here the most representative parser is Spell~\cite{du2016spell}.

Our method relates to the novel \textbf{Neural} category in the taxonomy of log parsing methods. Different from the current state-of-the-art heuristic-based methods, our method does not require any domain knowledge. Through empirical results, we show that the model is robust and applicable to a range of log types in different systems. We believe that in future this category will have the most influence considering the advances of deep learning. 

\section{Neural Log Parsing}\label{methodology}

\subsection{Preliminaries}\label{preliminaries}

We define the logs as sequences of temporally ordered unstructured text messages $L=(l_{i} \,:\,i=1,2,...)$, where each message $l_{i}$ is generated by a logging instruction (e.g. \textit{printf()}, \textit{log.info()}) within the software source code, and $i$ is its positional index within the sequence. 
The log messages consist of a constant and an optional varying part, respectively referred to as log template and variables. We define log templates and variables as tuples $EV=((e_i, v_i)\,:\,e \in \mathbb{E},\,i=1,2,...)$, where $\mathbb{E}$ is the finite set of all log event templates, $K=|\mathbb{E}|$ is the number of all unique templates and $v_i$ is a list of variables for the respectively associated template. They are associated with its original log message by the positional index $i$. 

The smallest inseparable singleton object within a log message is a token. Each log message consists of a bounded sequence of tokens, $\mathbf{t_i}=(t_{j}\,:\,t \in  \mathbb{T},\,j=1,2,...,|\mathbf{t_i}|)$, where $ \mathbb{T}$ is a set of all tokens, $j$ is the positional index of a token within the log message $l_i$, and $|\mathbf{t_i}|$ is the total number of tokens in $l_i$. For different $l_i$, $|\mathbf{t_i}|$ can vary. Depending on the concrete tokenization method, $t$ can be a word, word piece, or character. Therefore, tokenization is defined as a transformation function $\mathcal{T}: l_i \to \mathbf{t_i}, \forall i$.

With respect to our proposed log parsing method, the notions of context and embedding vector are additionally introduced. Given a token $t_j$, its context is defined by a preceding and subsequent sequence of tokens, i.e. a tuple of sequences: $C(t_j)=((t_{a}, t_{a+1},...,t_{j-1}), \allowbreak (t_{j+1}, t_{j+2},...,t_{b}))$, where $a<j<b$. An embedding vector is a $d$-dimensional real valued vector representation $\mathbf{s} \in \mathbb{R}^{d}$ of either a token or a log message.

We establish a requirement and a property for the proposed log parsing method: 
\textbf{Requirement 1} Given a temporally ordered sequence of log messages $L$, generated from an unknown set $\mathbb{E}$ of distinct log templates, the log parsing method should provide a mapping function $f_1: L \to EV$.

\textbf{Property 1} is a desirable feature of a log template extractor. While, each log template maps to a finite set of values, bounded with the number of unique log templates, this features allows for vector representation of a log hence opens a possibility for addressing various downstream tasks.


The generated vector representations should be closer embedding vectors for log messaged belonging to the same log template and distant embedding vectors for log messages belonging to distinct log templates. For example, the embedding vectors for \textit{"Took 10 seconds to create a VM"} and \textit{"Took 9 seconds to create a VM"} should have a small distance while vectors for \textit{"Took 9 seconds to create a VM"} and \textit{"Failed to create VM 3"} should be distant.

The goal of the proposed method is to mimic an operator's comprehension of logs. Given the task of identifying all event templates in a log, a reasonable approach is to pay close attention to parts that re-appear constantly and ignore parts that change frequently within a certain context (e.g. per log message). This can be modelled as a probability distribution for each token conditioned on its context, i.e. $P(t_j|C(t_j))$. Such probability distribution would allow the distinction of constant and varying tokens, referring to solving \textbf{Requirement 1}. The generation of log embedding vectors would naturally enable utilization of such representation for fine-tuning in downstream tasks. Moreover, the representation is obtained by focusing on constant parts of the log message, as they are more predictable, providing the necessary generalization for \textbf{Property 1}.

\subsection{NuLog: Self-Attentive Neural Parsing with Transformers}\label{NuLog}

\begin{figure}[!t]
\centerline{\includegraphics[width=\textwidth]{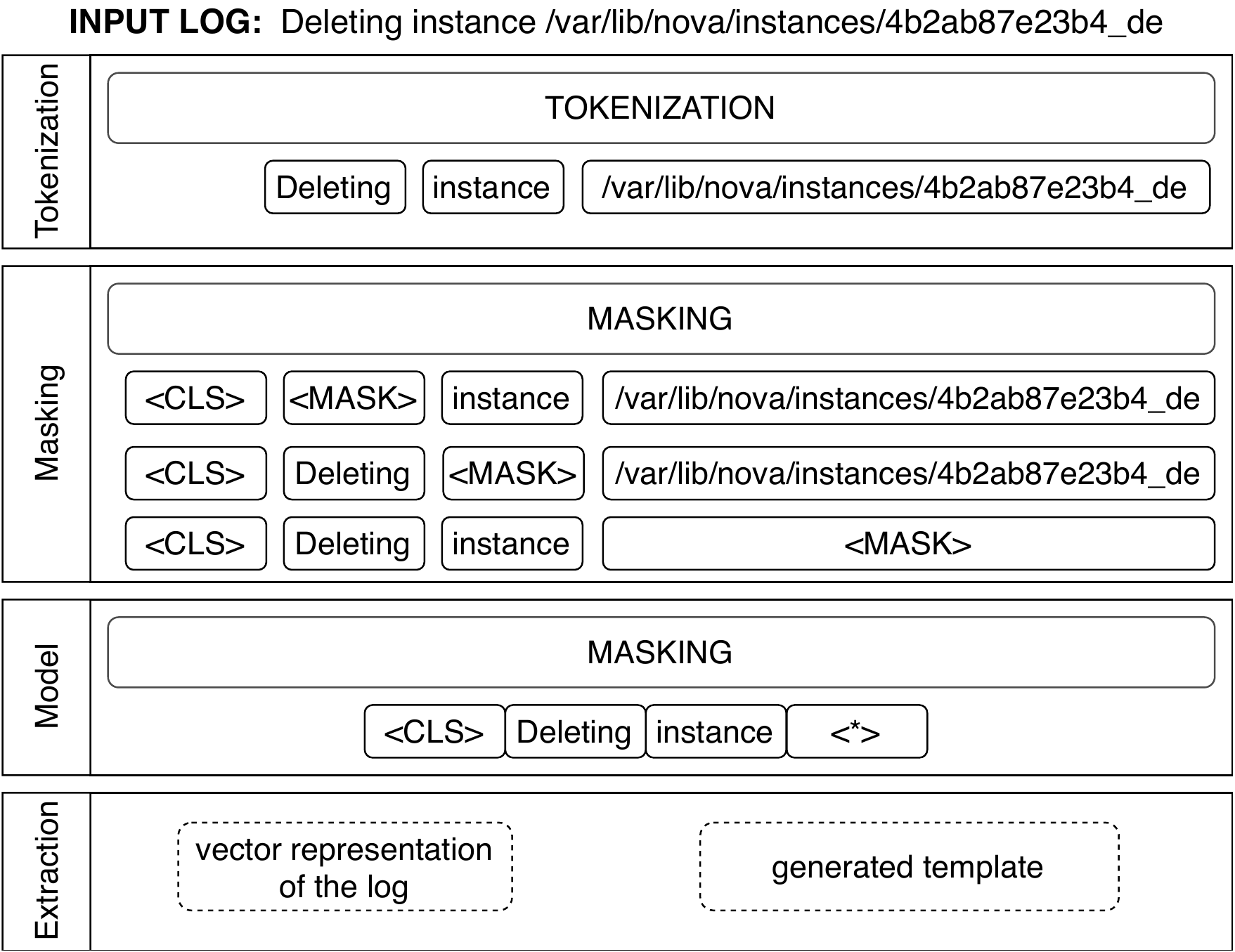}}
\caption{Instance of parsing of a single log message with \texttt{NuLog}.}
\label{fig:3}
\end{figure}

The proposed methods are composed of preprocessing, model, and template extraction. The overall architecture based on an example log message input is depicted in \figurename~\ref{fig:3}.

The log preprocessor transforms the log messages into a suitable format for the model. It is composed of two main parts: tokenization and masking. Before the tokenization task, the meta-information from the logging frameworks is stripped, and the payload, i.e., the print statement, is used as input to the tokenization step.

\textbf{Tokenization.} Tokenization transforms each log message into a sequence of tokens. For \texttt{NuLog}, we utilize a simple filter based splitting criterion to perform a string split operation. We keep these filters short and simple, i.e. easy to construct. All concrete criteria are described in section~\ref{Datasets}. In \figurename~\ref{fig:3} we illustrate the tokenization of the log message "\textit{Deleting instance /var/lib/nova/instances/4b2ab87e23b4de}". If a splitting criterion matches white spaces, then the log message is tokenized as a list of three tokens ["\textit{Deleting}", "\textit{instance}", "\textit{/var/lib/nova/instances/4b2ab87e23b4de}"]. In contrast to several related approaches that use additional hand-crafted regular expressions to parses parameters like IP addresses, numbers, and URLs, we do not parse any parameters with a regex expression. Such an approach is known to be error-prone and requires manual adjustments in different systems and even updates within the same system.

\textbf{Masking.} The intuition behind the proposed parsing method is to learn a general semantic representation of the log data by analyzing occurrences of tokens within their context. We apply a general method from natural language (NLP) research called Masked Language Modeling (MLM). It is originally introduced in~\cite{cloze} (where it is referred to as \textit{Cloze}) and successfully applied in other NLP publications like~\cite{devlin2018bert}. Our masking module takes the output of the tokenization step as input, which is a token sequence of a log message. A token from the sequence is randomly chosen and replaced with the special \texttt{$\langle MASK \rangle$} token. The masked token sequence is used as input for the model, while the masked token acts as the prediction target. To denote the start and end of a log message, we prepend a special \texttt{$\langle CLS \rangle$} and apply padding with \texttt{$\langle SPEC \rangle$} tokens. The number of padding tokens for each log message is given by $M - |\mathbf{t_i}|$, where $M=\max(|\mathbf{t_i}|) + 1, \; \forall i$ is the maximal number of tokens across all log messages within the log dataset added by one, and $|\mathbf{t_i}|$ is the number of tokens in the \textit{i-th} log message. Note, that the added one ensures that each log message is padded by at least one \texttt{$\langle SPEC \rangle$} token.

\textbf{Model}. The method has two operation modes - offline and online. During the offline phase, log messages are used to tune all model parameters via backpropagation and optimal hyper-parameters are selected. During the online phase, every log message is passed forward through the model. This generates the respective log template and an embedding vector for each log message.

\begin{figure}[htbp]
\centerline{\includegraphics[width=\textwidth]{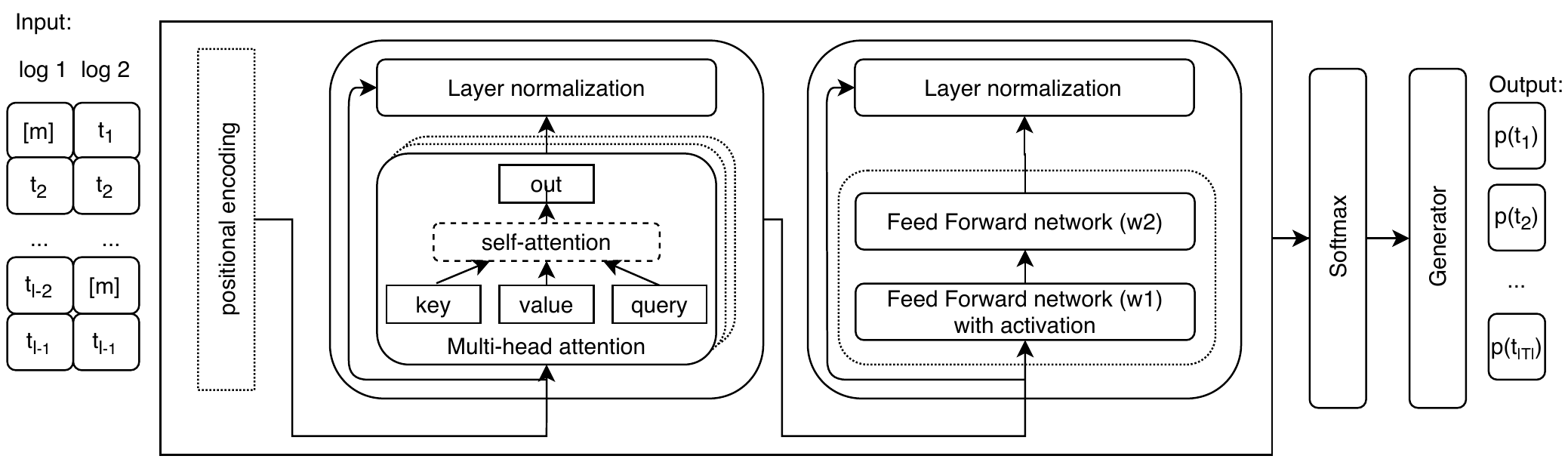}}
\caption{Model architecture of \texttt{NuLog} for parsing of the logs.}
\label{fig:2}
\end{figure}

\figurename~\ref{fig:2} depicts the complete architecture. The model applies two operations on the input token vectors: token vectorization and positional encoding. The subsequent encoder structure takes the result of these operations as input. It is composed of two elements: self-attention layer and feedforward layer. The last model component is a single linear layer with a softmax activation overall tokens appearing in the logs. In the following, we provide a detailed explanation of each model element.

Since all subsequent elements of the model expect numerical inputs, we initially transform the tokens into randomly initialized numerical vectors $\mathbf{x} \in \mathbb{R}^d$. These vectors are referred to as token embeddings and are part of the training process, which means they are adjusted during training to represent the semantic meaning of tokens depending on their context. These numerical token embeddings are passed to the positional encoding block. In contrast to e.g., recurrent architectures,  attention-based models do not contain any notion of input order. Therefore, this information needs to be explicitly encoded and merged with the input vectors to take their position within the log message into account. This block calculates a vector $\mathbf{p} \in \mathbb{R}^d$ representing the relative position of a token based on a sine and cosine function.

\begin{equation}
    p_{2k}=sin \left( \frac{j}{10000^{\frac{2k}{v}}} \right), \;\;
    p_{2k+1}=cos \left( \frac{j}{10000^{\frac{2k + 1}{v}}} \right).
\end{equation}

Here, $k=0,1,\dots,d-1$ is the index of each element in $\mathbf{p}$ and $j=0,1,\dots,M$ is the positional index of each token. Within the equations, the parameter $k$ describes an exponential relationship between each value of vector $\mathbf{p}$. Additionally, a sine and cosine function are interchangeably applied. Both allow better discrimination of the respective values within a specific vector of $\mathbf{p}$. Furthermore, both functions have an approximately linear dependence on the position parameter $j$, which is hypothesized to make it easy for the model to attend to the respective positions. Finally, both vectors can be combined as $\mathbf{x'} = \mathbf{x} + \mathbf{p}$.


The encoder block of our model starts with a multi-head attention element, where a softmax distribution over the token embeddings is calculated. Intuitively, it describes the significance of each embedding vector for the prediction of the target masked token. We summarize all token embedding vectors as rows of a matrix $X'$ and apply the following formula
\begin{equation}
    Z_l=softmax \left( \frac{Q_l \times K^T_l}{\sqrt{w}} \right)V_l, \; \text{for} \; l = 1, 2, \dots, L,
\end{equation}
where $L$ denotes the number of attention heads, $w = \frac{d}{L}$ and $d \, \text{mod} \, L = 0$. The parameters $Q$, $K$ and $V$ are matrices, that correspond to the query, key, and value elements in \figurename~\ref{fig:2}. They are obtained by applying matrix multiplications between the input $X'$ and respective learnable weight matrices $W_{l}^{Q}$, $W_{l}^{K}$, $W_{l}^{V}$:

\begin{equation}
    Q_l= X^* \times W_{l}^{Q}, \; K_l= X^* \times W_{l}^{K}, \; V_l= X^* \times W_{l}^{V},
\end{equation}

where $W_{l}^{Q}, \; W_{l}^{K}, \; W_{l}^{V} \in \mathbb{R}^{M \times w}$. The division by $\sqrt{w}$ stabilizes the gradients during training. After that, the softmax function is applied and the result is used to scale each token embedding vector: $X''_l=X'_l \times Z_l$. The scaled matrices $X''_l$ are concatenated to a single matrix $X''$ of size $M \times d$. As depicted in \figurename~\ref{fig:2} there is a residual connection between the input token matrix $X'$ and its respective attention transformation $X''$, followed by a normalization layer $norm$. These are used for improving the performance of the model by tackling different potential problems encountered during the learning such as small gradients and the covariate shift phenomena. Based on this, the original input is updated by the attention-transformed equivalent as $X' = norm(X' + X'')$.

The last element of the encoder consists of two feed-forward linear layers with a ReLU activation in between. It is applied individually on each row of $X'$. Thereby, identical weights for every row are used, which can be described as a convolution over each attention-transformed matrix row with kernel size one. This step serves as additional information enrichment for the embeddings. Again, a residual connection followed by a normalization layer between the input matrix and the output of both layers is employed. This model element preserves the dimensionality $X'$.

The final element of the model consists of a single linear layer. It receives the encoder result $X'$ and extracts the token embedding vector of the \texttt{$\langle CLS \rangle$} token. Since every log message token sequence is pre-padded by this special token, it is the first row of the matrix, i.e. $\mathbf{x'_0} \in X'$. The linear layer maps this vector of size $d$ to a vector whose size corresponds to the total number of tokens $|\mathbb{T}|$ in the dataset. The subsequent softmax is utilized to calculate a probability distribution over each element of $\mathbb{T}$. During training, the masked token is used as the target to be predicted. Since the last vector embedding of the \texttt{$\langle CLS \rangle$} token is used for prediction, it is forced to summarize the log message. Otherwise, it would not be able to solve the masked token prediction task well enough across all tokens. We hypothesize that the constant part of log templates will constraint the model to learn similar \texttt{$\langle CLS \rangle$} token embeddings when log messages are of the same template. This leads to a mapping of the log messages to their vector representation, which can after be used for diverse downstream tasks like anomaly detection. This log message embedding vector satisfies the proposed \textbf{Property 1} (see Section~\ref{preliminaries}).

\subsection{Log Template Extraction}
The extraction of all log templates within a log dataset is executed online, after the model training. Therefore, we pass each log message as input and configure the masking module in a way that every token is masked consecutively, one at a time. We measure the model's ability to predict each token, and thus, decide whether the token is a constant part of the template or a variable. High confidence in the prediction of a specific token indicates a constant part of the template, while small confidence is interpreted as a variable. More specifically, we employ the following procedure. If the prediction of a particular token is in the top $\epsilon$ predictions, we consider it to be part of the constant part of the template, otherwise, it is considered to be a variable. For all variables, an indicator \texttt{$\langle * \rangle$} is placed on its position within the log message. 
This addresses the \textbf{Requirement 1} proposed in Section~\ref{preliminaries}.


\section{Evaluation}\label{evaluation}
To quantify the performance of the proposed method, we perform an exhaustive evaluation of the log parsing task on a set of ten benchmark datasets and compare the results with twelve other log template parsing methods. The datasets together with the implementation of the other parsers were obtained from the log benchmark~\cite{zhu2019tools}. Furthermore, the model of \texttt{NuLog} provides log message vector embeddings. We show that these, along with the model, can be used for anomaly detection as downstream tasks.

\begin{table}[htbp]
\centering
\caption{Log datasets and the number of log templates.}
\begin{tabular}{llc}
\hline
System    & Description                                                                                  & \#Templates \\ \hline \hline
BGL       & \begin{tabular}[c]{@{}l@{}}BlueGene Supercomputer\end{tabular}                              & 120         \\
Android   & \begin{tabular}[c]{@{}l@{}}Mobile Operating System\end{tabular}                             & 166         \\
OpenStack & \begin{tabular}[c]{@{}l@{}}Cloud Operating System\end{tabular}                             & 43          \\
HDFS      & \begin{tabular}[c]{@{}l@{}}Hadoop Distributed File System\end{tabular}                    & 14          \\
Apache    & \begin{tabular}[c]{@{}l@{}}Apache HTTP Server\end{tabular}                                & 6           \\
HPC       & \begin{tabular}[c]{@{}l@{}}High Performance Cluster (Los Alamos)\end{tabular}            & 46          \\
Windows   & \begin{tabular}[c]{@{}l@{}}Windows 7 Computer Operating System\end{tabular}               & 50          \\
HealthApp & \begin{tabular}[c]{@{}l@{}}Mobile Application for Andriod Devices\end{tabular}             & 75          \\
Mac       & \begin{tabular}[c]{@{}l@{}}MacOS Operating System\end{tabular}                            & 341         \\
Spark     & \begin{tabular}[c]{@{}l@{}}Unified Analytics Engine for Big Data Processing,\end{tabular} & 36          \\ \hline
\end{tabular}
\label{table:logdatasets}
\end{table}

\subsection{Datasets}\label{Datasets}
The log datasets employed in our experiments are summarized in Table~\ref{table:logdatasets}. These real-world log data range from supercomputer logs (BGL and HPC), distributed system logs (HDFS, OpenStack, Spark), to standalone software logs (Apache, Windows, Mac, Android). To enable reproducibility, we follow the guidelines from~\cite{zhu2019tools} and utilize a random sample of 2000 log messages from each dataset, where the ground truth templates are available. The number of templates contained within each dataset is shown in table \ref{table:logdatasets}.  

The BGL dataset is collected by Lawrence Livermore National Labs (LLNL) from BlueGene/L supercomputer system. HPC logs are collected from a high-performance cluster, consisting of 49 nodes with 6,152 cores. HDFS is a log data set collected from the Hadoop distributed file system deployed on a cluster of 203 nodes within the Amazon EC2 platform. OpenStack is a result of a conducted anomaly experiment within CloudLab with one control node, one network node and eight compute nodes. Spark is an aggregation of logs from the Spark system deployed within the Chinese University of Hongkong, which comprises 32 machines. The Apache HTTP server dataset consists of access and error logs from the apache web server. Windows, Mac, and Android datasets consist of logs generated from single machines using the respectively named operating system. HealthApp contains logs from an Android health application, recorded over ten days on a single android smartphone.

As described in Section~\ref{NuLog}, the tokenization process of our method is implemented by splitting based on a filter. We list the applied splitting expressions for each dataset in Table~\ref{table:hyperparameters}. Besides, we also list the additional training parameters. The number of epochs is determined by an early stopping criterion, which terminated the learning when the loss converges. The hyperparameter $\epsilon$ is determined via cross-validation.

\begin{table}[!t]
\centering
\caption{\texttt{NuLog} hyperparameter setting.}
\begin{tabular}{llcc}
\hline
System    & Tokenization filter & \#epochs & $\epsilon$   \\ \hline \hline
BGL       & \Verb #([ |:|\(|\)|=|,])|(core.)|(\.{2,})# & 3        & 50  \\
Android   & \Verb #([ |:|\(|\)|=|,|"|\{|\}|@|\$|\[|\]|\||;])# & 5        & 25  \\
OpenStack & \Verb #([ |:|\(|\)|"|\{|\}|@|\$|\[|\]|\||;])# & 6        & 5   \\
HDFS      & \Verb #(\s+blk_)|(:)|(\s)# & 5        & 15  \\
Apache    & \Verb #([ ])# & 5        & 12  \\
HPC       & \Verb #([ |=])# & 3        & 10  \\
Windows   & \Verb #([ ])# & 5        & 95  \\
HealthApp & \Verb #([ ])# & 5        & 100 \\
Mac       & \Verb #([ ])|([\w-]+\.){2,}[\w-]+# & 10       & 300 \\
Spark     & \Verb #([ ])|(\d+\sB)|(\d+\sKB)|(\d+\.){3}\d+# & 3        & 50  \\ \hline
\end{tabular}
\label{table:hyperparameters}
\end{table}

\subsection{Evaluation methods}\label{ComparisonMethods}
To quantify the effectiveness of \texttt{NuLog} for log template generation from the presented eleven datasets, we compare it with twelve existing log parsing methods on parsing accuracy, edit distance, and robustness. We reproduced the results from Zhu et al.~\cite{zhu2019tools} for all known log parsers. Furthermore, we enriched the extensive benchmark reported by an additional metric, i.e., edit distance. Note, that all methods we comparing with are described in detail in Section~\ref{relatedwork}. To evaluate the log message embeddings for the anomaly detection downstream tasks, we use the common metrics accuracy, recall, precision, and F1 score. In the following, we describe each evaluation metric.

\textbf{Parsing Accuracy}. To enable comparability between our method to the ones analyzed in the benchmark~\cite{zhu2019tools}, we adopt their proposed parsing accuracy ($PA$) metric. It is defined as the ratio of correctly parsed log messages over the total number of log messages. After parsing, each log message is assigned to a log template. A log message is considered correctly parsed if its log template corresponds to the same group of log messages as the ground truth does. For example, if a log sequence $[e_1, e_2, e_2]$ is parsed to $[e_1, e_4, e_5]$, we get $PA=\frac{1}{3}$ since the second and third messages are not grouped together.

\textbf{Edit distance}.
The $PA$ metric is considered as the standard for evaluation of log parsing methods, but it has limitations when it comes to evaluating the template extraction in terms of string comparison. Consider a particular group of logs produced from single \textit{print("VM created successfully")} statement that is parsed with the word \textit{Template}. As long as this is consistent over every occurrence of the templates from this group throughout the dataset, $PA$ would still yield a perfect score for this template parsing result, regardless of the obvious error. Therefore, we introduce an additional evaluation metric: Levenshtein edit distance. This is a way of quantifying how dissimilar two log messages are to one another by counting the minimum number of operations required to transform one message into the other.


\subsection{Parsing Results}
\subsubsection{Parsing Accuracy}

This section presents and discusses the log parsing $PA$ results of \texttt{NuLog} on the benchmark datasets and compares them with twelve other related methods. These are presented in table~\ref{table:comparisonPA}. Specifically, each row contains the datasets while the compared methods are represented in the table columns. Additionally, the penultimate column contains the highest value of the first twelve columns - referred to as best of all - and the last column contains the results for \texttt{NuLog}. In the bold text, we highlight the best of the methods per dataset. HDFS and Apache datasets are most frequently parsed with 100\% $PA$. This is because HDFS and Apache error logs have relatively unambiguous event templates that are simple to identify. On those, \texttt{NuLog} achieves comparable results. For the Spark, BGL and Windows dataset, the existing methods already achieve high $PA$ values above 96\% (BGL) or above 99\% (Spark and Windows). Our proposed method can slightly outperform those. For the rather complex log data from OpenStack, HPC and HealthApp the baseline methods achieve a $PA$ between 78\% and 90\%, which \texttt{NuLog} significantly outperforms by 4-13\%.
\begin{table}[htbp]
\centering
\caption{Comparisons of log parsers and our method \texttt{NuLog} in parsing accuracy (PA).}
\resizebox{\columnwidth}{!}{%
\begin{tabular}{|l|cccccccccccccc|}
\hline
Dataset   & SLCT           & AEL            & LKE   & LFA            & LogSig & SHISHO & LogCluster & LenMa & LogMine        & Spell          & Drain          & MoLFI & BoA          & NuLog           \\ \hline \hline
HDFS      & 0.545          & 0.998          & 1.000 & 0.885          & 0.850  & 0.998  & 0.546      & 0.998 & 0.851          & 1.000          & 0.998          & 0.998 & \textbf{1.000} & 0.998          \\
Spark     & 0.685          & 0.905          & 0.634 & \textbf{0.994} & 0.544  & 0.906  & 0.799      & 0.884 & 0.576          & 0.905          & 0.920          & 0.418 & 0.994          & \textbf{1.000} \\
OpenStack & \textbf{0.867} & 0.758          & 0.787 & 0.200          & 0.200  & 0.722  & 0.696      & 0.743 & 0.743          & 0.764          & 0.733          & 0.213 & 0.867          & \textbf{0.990} \\
BGL       & 0.573          & 0.758          & 0.128 & 0.854          & 0.227  & 0.711  & 0.835      & 0.690 & 0.723          & 0.787          & \textbf{0.963} & 0.960 & 0.963          & \textbf{0.980} \\
HPC       & 0.839          & \textbf{0.903} & 0.574 & 0.817          & 0.354  & 0.325  & 0.788      & 0.830 & 0.784          & 0.654          & 0.887          & 0.824 & 0.903          & \textbf{0.945} \\
Windows   & 0.697          & 0.690          & 0.990 & 0.588          & 0.689  & 0.701  & 0.713      & 0.566 & 0.993          & 0.989          & \textbf{0.997} & 0.406 & 0.997          & \textbf{0.998} \\
Mac       & 0.558          & 0.764          & 0.369 & 0.599          & 0.478  & 0.595  & 0.604      & 0.698 & \textbf{0.872} & 0.757          & 0.787          & 0.636 & \textbf{0.872} & 0.821          \\
Android   & 0.882          & 0.682          & 0.909 & 0.616          & 0.548  & 0.585  & 0.798      & 0.880 & 0.504          & \textbf{0.919} & 0.911          & 0.788 & \textbf{0.919} & 0.827          \\
HealthApp & 0.331          & 0.568          & 0.592 & 0.549          & 0.235  & 0.397  & 0.531      & 0.174 & 0.684          & 0.639          & \textbf{0.780} & 0.440 & 0.780          & \textbf{0.875} \\
Apache    & 0.731          & 1.000          & 1.000 & 1.000          & 1.000  & 1.000  & 0.709      & 1.000 & 1.000          & 1.000          & 1.000          & 1.000 & \textbf{1.000} & \textbf{1.000} \\ \hline
\end{tabular}
}
\label{table:comparisonPA}
\end{table}
\subsubsection{PA robustness}
\begin{figure}[htbp]
\centerline{\includegraphics[scale=0.32]{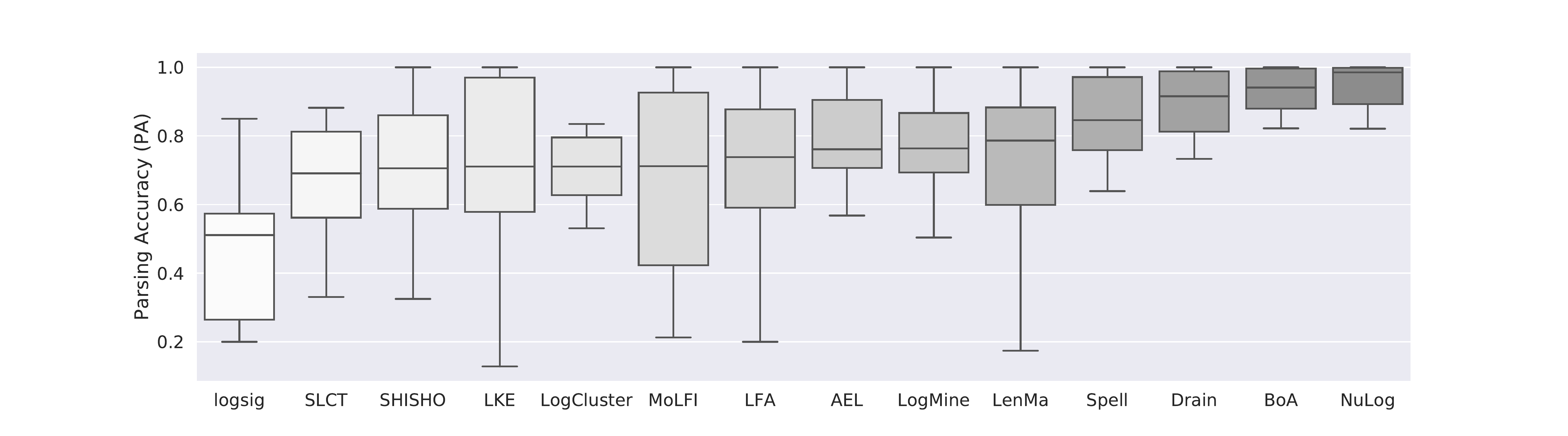}}
\caption{Robustness evaluation on the parsing accuracy of the log parsers.}
\label{robustness_pa}
\end{figure}

Employing a general parsing method in production requires a robust performance throughout different log datasets. With the proposed method, we explicitly aim at supporting a broad range of diverse log data types. Therefore, the robustness of \texttt{NuLog} is analyzed and compared to the related methods. \figurename~\ref{robustness_pa} shows the accuracy distribution of each log parser across the log datasets within a boxplot. From left to right in the figure, the log parsers are arranged in ascending order of the median $PA$. That is, LogSig has the lowest and \texttt{NuLog} obtains the highest parsing accuracy on the median.  We postulate the criterion of achieving consistently high $PA$ values across many different log types as crucial for their general use. However, it can be observed that, although most log parsing methods achieve high $PA$ values of 90\% for specific log datasets, they have a large variance when applied across all given log types. \texttt{NuLog} outperforms every other baseline method in terms of $PA$ robustness yielding a median of 0.99, which even lies above the best of all median of 0.94.

\subsubsection{Edit distance}

As an evaluation metric, $PA$ measures how well the parsing method can match log templates with the respective log messages throughout the dataset. Additionally, we want to verify the correctness of the templates, e.g., whether all variables are correctly identified. To achieve this, the edit distance score is employed to measure the dissimilarity between the parsed and the ground truth log templates. Note that this indicates that the objective is to achieve low edit distance values. All edit distance scores are listed in table~\ref{table:edit_distance}. The table structure is the same as for $PA$ results. In bold we highlight the best edit distance value across all tested methods per dataset. It can be seen that in terms of edit distance \texttt{NuLog} outperforms existing methods on the HDFS, Windows, Android, HealthApp and Mac datasets. It performs comparable on the BGL, HPC, Apache and OpenStack datasets and achieves a higher edit distance on the Spark log data.

\subsubsection{Edit distance robustness}
\begin{figure}[htbp]
\centerline{\includegraphics[scale=0.32]{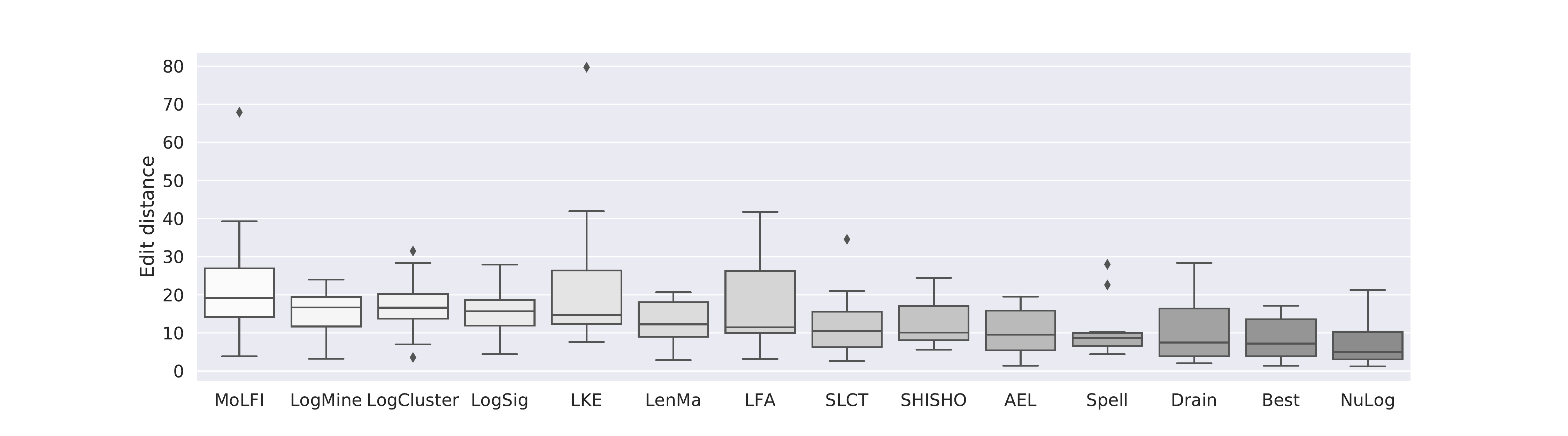}}
\caption{Robustness evaluation on the edit distance of the log parsers.}
\label{robustness_ed}
\end{figure}

Similar to the $PA$ robustness evaluation, we want to verify how consistent \texttt{NuLog} is performing in terms of edit distance across the different log datasets. \figurename~\ref{robustness_ed} shows a box-plot that indicates the edit distance distribution of each log parser for all log datasets. From left to right in the figure, the log parsing methods are arranged in descending order of the median edit distance. Again, it can be observed that although most log parsing methods achieve the minimal edit distance scores under 10, most of them have a large variance over different datasets and are therefore not generally applicable for diverse log data types. MoLFI has the highest median edit distance, while Spell and Drain perform constantly well - i.e. small median edit distance values - for multiple datasets. Again, our proposed parsing method outperforms the lowest edit distance values with a median of 5.00, which is smaller the best of all median of 7.22.

\begin{table}[htbp]
\centering
\caption{Comparisons of log parsers and our method \texttt{NuLog} in edit distance.}
\resizebox{\columnwidth}{!}{%
\begin{tabular}{|l|cccccccccccccc|}
\hline
Dataset &  LogSig &      LKE &    MoLFI &     SLCT &      LFA &  LogCluster &   SHISHO &  LogMine &    LenMa &    Spell &      AEL &    Drain  &  BoA&    NuLog \\
\hline \hline
HDFS & 19.1595 &  17.9405 &  19.8430 &  13.6410 &  30.8190 &     28.3405 &  10.1145 &  16.2495 &  10.7620 &   9.2740 &   8.8200 &   8.8195 &    8.8195& \textbf{3.2040} \\
Spark & 13.0615 &  41.9175 &  14.1880 &   6.0275 &   9.1785 &     17.0820 &   7.9100 &  16.0040 &  10.9450 &   6.1290 &   3.8610 &   \textbf{3.5325} &  3.5325&  12.0800 \\
BGL &  11.5420 &  12.5820 &  10.9250 &   9.8410 &  12.5240 &     12.9550 &   8.6305 &  19.2710 &   8.3730 &   7.9005 &   5.0140 &   \textbf{4.9295} &   4.9295&  5.5230 \\
HPC &  4.4475 &   7.6490 &   3.8710 &   2.6250 &   3.1825 &      3.5795 &   7.8535 &   3.2185 &   2.9055 &   5.1290 &   \textbf{1.4050} &   2.0155 &   1.4050&      2.9595 \\
Windows &  7.6645 &  11.8335 &  14.1630 &   7.0065 &  10.2385 &      6.9670 &   5.6245 &   6.9190 &  20.6615 &   4.4055 &  11.9750 &   6.1720 &   5.6245&     \textbf{4.4860} \\
Android & 16.9295 &  12.3505 &  39.2700 &   3.7580 &   9.9980 &     16.4175 &  10.1505 &  22.5325 &   3.2555 &   8.6680 &   6.6550 &   3.2210 &    3.2210&    \textbf{1.1905} \\
HealthApp & 17.1120 &  14.6675 &  21.6485 &  16.2365 &  20.2740 &     16.8455 &  24.4310 &  19.5045 &  16.5390 &   8.5345 &  19.0870 &  18.4965 &    14.6675&  \textbf{6.2075} \\
Apache & 14.4420 &  14.7115 &  18.4410 &  11.0260 &  10.3675 &     16.2765 &  12.4405 &  10.2655 &  13.5520 &  10.2335 &  10.2175 &  \textbf{10.2175} &   10.2175& 11.6915 \\
OpenStack & 21.8810 &  29.1730 &  67.8850 &  20.9855 &  28.1385 &     31.4860 &  18.5820 &  23.9795 &  18.5350 &  27.9840 &  \textbf{17.1425} &  28.3855 &   17.1425&  21.2605 \\
Mac & 27.9230 &  79.6790 &  28.7160 &  34.5600 &  41.8040 &     21.3275 &  19.8105 &  17.0620 &  19.9835 &  22.5930 &  19.5340 &  19.8815 &      17.062& \textbf{2.8920} \\
\hline
\end{tabular}
}

\label{table:edit_distance}
\end{table}

\section{Case study: Anomaly detection as a downstream task}
Anomaly detection in complex and distributed systems is a crucial task in distributed and complex IT systems. The on-time detection provides a way to take action towards preventing or fast-reacting to emerging problems. Ultimately, it allows the operator to satisfy the service level agreements.

Our model architecture allows for coupling of the parsing approach and a downstream anomaly detection task. The knowledge obtained during the log parsing phase is used as a good prior bias for the downstream task. The architecture provides treating the problem of anomaly detection in both the supervised and unsupervised way. To illustrate this we designed two experimental case studies described in the following.

\begin{figure}[!htbp]
\centerline{\includegraphics[width=0.99\textwidth]{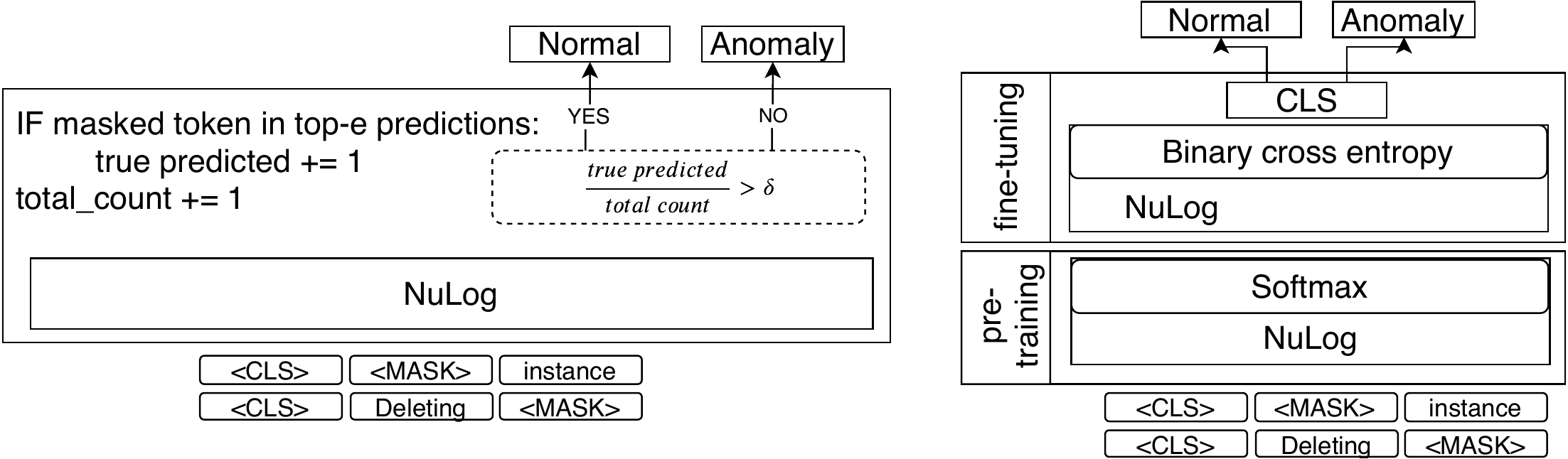}}
\caption{Unsupervised (left) and supervised (supervised) methods for downstream anomaly detection.}
\label{fig:downstream}
\end{figure}

\subsection{Unsupervised anomaly detection}
We test the log message embedding produced by \texttt{NuLog} for unsupervised log anomaly detection by employing a similar approach as during the parsing. We train the model for three epochs. Each token of a log message is masked and predicted based on the \texttt{$\langle CLS \rangle$} token embedding. All respectively masked tokens that are not in the top-$\epsilon$ predicted tokens are marked as anomalies. We compute the percentage of anomalous tokens within the log message to decide whether the whole log message is anomalous. If it is larger than a threshold $\delta$, the log message is considered as an anomaly, otherwise as normal. We show this process in the left part of \figurename~\ref{fig:downstream}.

To the best of our knowledge, only the BGL dataset contains anomaly labels for each individual log message, and is, therefore, suitable to evaluate the proposed anomaly detection approach. Due to its large volume, we use only the first 10\% of it. For training 80\% of that portion is utilized, while the rest is used for testing. In the first row of table~\ref{table:anomalydetection} we show the accuracy, recall, precision, and F1 score results. It can be seen that the method yields scores between 0.999 and 1.0. We, therefore, regard these results as evidence that the log message embeddings can be used for the unsupervised detection of anomalous log messages.

\begin{table}[!htbp]
\centering
\caption{Scores for the downstream anomaly detection tasks.}
\begin{tabular}{|l|c|c|c|c|}
\hline
             & Accuracy & Recall & Precision & F1 Score \\ \hline
Unsupervised & 0.999    & 0.999  & 1.000     & 0.999    \\
 \hline
Supervised   & 0.999    & 1.000  & 0.999     & 0.999    \\
\hline
\end{tabular}
\label{table:anomalydetection}
\end{table}

\subsection{Supervised anomaly detection}
For the second case study, we utilize log message embedding as a feature for supervised anomaly detection. The model is first trained on the self-supervised MLM task. After that, we replace the last softmax layer by a linear layer, that is adapted via supervised training of predicting a given \texttt{$\langle CLS \rangle$} as either normal or anomaly, i.e., binary classification. For this downstream task, we applied a fine-tuning of two epochs.

The first 10\% of the BGL dataset were used for evaluation. Thereby, the model is trained on the first 80\% and evaluated on the remaining 20\%. The results are listed in the second row of Table~\ref{table:anomalydetection} and show that two epochs of fine-tuning are sufficient to produce an F1 score of 0.99. It further adds evidence to the proposed hypothesize of enabling the application of the semantic log message embedding for different downstream tasks.

\section{Conclusion} \label{conclusion}
To address the problem of log parsing we adopt the masked word prediction learning task. The insight of having words appearing on the constant position of the log entry means that their correct prediction directly produces the log message type. The incorrect token prediction reflects various parts of the logs as are its parameters. The method also produces a numerical representation of the context of the log message, which primarily is utilized for parsing. This allows the model for utilization in downstream tasks such as anomaly detection.

To evaluate the effectiveness of \texttt{NuLog}, we conducted experiments on 10 real-world log datasets and evaluated it against 12 log parsers. Furthermore, we enhanced the evaluation protocol with the addition of a new measure to justify the offset of generated templates and the true log message types. The experimental results show that \texttt{NuLog} outperforms the existing log parsers in terms of
accuracy, edit distance, and robustness. Furthermore, we conducted case studies on a real-world supervised and unsupervised anomaly detection task. The results show that the model and the representation learned during parsing with masked language modeling are beneficial for distinguishing between normal and abnormal logs in both supervised and unsupervised scenario.

Our approach shows that log parsing can be performed with deep language modeling. This imply that future research in log parsing and anomaly detection should focus more into generalization accross domains, transfer of knowledge, and learning of meaningful log representations that could further improve the troubleshooting tasks critical for operation of IT systems.
\bibliographystyle{splncs04}
\bibliography{lncsmain}

\end{document}